\PassOptionsToPackage{numbers}{natbib}
\documentclass{article}
\usepackage[preprint]{neurips_2026}

\usepackage[utf8]{inputenc}
\usepackage[T1]{fontenc}
\usepackage{hyperref}
\usepackage{url}
\usepackage{booktabs}
\usepackage{amsfonts}
\usepackage{amsmath}
\usepackage{amssymb}
\usepackage{nicefrac}
\usepackage{microtype}
\usepackage{xcolor}
\usepackage{graphicx}
\usepackage{subcaption}
\usepackage{multirow}
\usepackage{array}
\usepackage{float}
\usepackage{colortbl}

\definecolor{bestrow}{RGB}{230, 242, 255}

\title{DroneScan-YOLO: Redundancy-Aware Lightweight Detection\\
       for Tiny Objects in UAV Imagery}

\author{
  Yann V. Bellec \\
  \texttt{ybellec@ucsd.edu}
}

\begin{document}
\maketitle

\begin{abstract}
Aerial object detection in UAV imagery presents unique challenges due to the high
prevalence of tiny objects, adverse environmental conditions, and strict computational
constraints. Standard YOLO-based detectors fail to address these jointly: their minimum
detection stride of 8 pixels renders sub-32px objects nearly undetectable, their CIoU ~\cite{ciou}
loss produces zero gradients for non-overlapping tiny boxes, and their architectures
contain significant filter redundancy. We propose \textbf{DroneScan-YOLO}, a holistic
system contribution that addresses these limitations through four coordinated design
choices: (1)~increased input resolution of $1280{\times}1280$ to maximize spatial detail
for tiny objects, (2)~\textbf{RPA-Block}, a dynamic filter pruning mechanism based on
lazy cosine-similarity updates with a 10-epoch warm-up period, (3)~\textbf{MSFD}, a
lightweight P2 detection branch at stride~4 adding only 114{,}592 parameters (+1.1\%),
and (4)~\textbf{SAL-NWD}, a hybrid loss combining Normalized Wasserstein Distance with
size-adaptive CIoU weighting, integrated into YOLOv8's TaskAligned assignment pipeline.
Evaluated on VisDrone2019-DET, DroneScan-YOLO achieves \textbf{55.3\%} mAP@50 and
\textbf{35.6\%} mAP@50-95, outperforming the YOLOv8s baseline by $+$16.6 and $+$12.3
points respectively, improving recall from 0.374 to 0.518, and maintaining 96.7~FPS
inference speed with only $+$4.1\% parameters. Gains are most pronounced on tiny object
classes: bicycle AP@50 improves from 0.114 to 0.328 ($+$187\%), and awning-tricycle
from 0.156 to 0.237 ($+$52\%).
\end{abstract}

\section{Introduction}

Drones and Unmanned Aerial Vehicles (UAVs) take an increasingly considerable place in
industry, where their use is found for surveillance, traffic monitoring, search and
rescue, agricultural inspection, and border security. The performance of real-time
object detection from a UAV camera is a major issue for these applications. Despite
numerous significant advances in deep learning, certain important limitations persist
in UAV imaging. We have selected and sought to resolve three fundamental and
interdependent challenges.

\textbf{Tiny object prevalence.}
Among camera data recorded at altitude, a major part of the targets occupy too few
pixels on the recording. In the VisDrone2019-DET dataset, 68\% of annotated instances
occupy less than $32{\times}32$ pixels. The YOLOv8s architecture generates predictions
at strides 8, 16 and 32, characterizing feature maps P3, P4 and P5. For stride~8,
an element of $8{\times}8$ pixels in a $640{\times}640$ image generates a $1{\times}1$
activation on P3, which results in a vector unfit to capture an exploitable context
for classification.

\textbf{Loss instability for small bounding boxes.}
Loss functions based on IoU and its variants (GIoU, DIoU, CIoU) demonstrate a notable
failure mode for small objects: the gradient cancels when predicted and target boxes do
not overlap. For a target box of $5{\times}5$ pixels, a shift of 1 pixel makes the IoU
fall from 0.25 to 0, suppressing the global learning signal for the most challenging
examples.

\textbf{Computational efficiency.}
Embedded UAV systems operate under strict constraints of power, memory and thermal
dissipation. Existing solutions such as adding detection heads or deeper backbones
improve precision at the direct cost of computational load and model size.

The current literature addresses these problems individually. YOLO-LE demonstrates
increased architectural efficiency but no improvement of the loss. DAU-YOLO improves
features by attention but does not address resolution or loss. NWD provides a robust
metric for small objects but does not fit into a synergistic framework. No method
explicitly combines input resolution, multi-scale detection, filter pruning and loss
function in a common operationalizable base.

We introduce \textbf{DroneScan-YOLO}, which addresses these constraints simultaneously:
\begin{enumerate}
  \item[\textbf{(I)}]   Increased resolution to 1280px, quadrupling feature map surface
        and improving tiny object detectability.
  \item[\textbf{(II)}]  \textbf{RPA-Block}: pruning system based on cosine similarity
        between filters, with warm-up period and lazy update to reduce computational
        redundancy.
  \item[\textbf{(III)}] \textbf{MSFD}: lightweight P2 detection branch at stride~4
        using separable convolutions and squeeze-and-excitation attention, adding
        only 114{,}592 parameters ($+$1.1\%).
  \item[\textbf{(IV)}]  \textbf{SAL-NWD}: hybrid loss associating Normalized Wasserstein
        Distance with inverse-area CIoU weighting, implemented in YOLOv8's
        TaskAligned pipeline.
  \item[\textbf{(V)}]   Comprehensive experimental evaluation on VisDrone2019-DET
        with ablation study on 8 configurations and sensitivity analyses for key
        hyperparameters.
\end{enumerate}

The remainder of this paper is organized as follows. Section~\ref{sec:related} reviews
related work. Section~\ref{sec:method} describes the proposed architecture.
Section~\ref{sec:experiments} presents experimental results.
Section~\ref{sec:conclusion} concludes.

\section{Related Work}
\label{sec:related}

Object detection by deep learning has known a long trajectory of progress since the
first two-stage architectures ~\cite{fasterrcnn} like R-CNN and its variants, towards the real-time
monolithic detectors of the YOLO family. YOLOv1 introduced the paradigm of single-pass
detection in 2016 ~\cite{yolov1}, followed by YOLOv3 ~\cite{yolov3} which established multi-scale detection via
predictions at three distinct resolutions. YOLOv5 consolidated this approach with a
highly optimized training pipeline. YOLOv8, developed by Ultralytics in 2023,
represents the state of the art of this family with a decoupled detection head, an
anchor-free design and an optimized CSP backbone, reaching competitive performances on
COCO at high inference speed. Nevertheless, its minimal detection stride of 8 pixels
constitutes an important constraint for UAV scenes where objects are of very small size.

\textbf{YOLO-based detectors for UAV imagery.}
A large number of studies aim to adapt YOLO architectures for UAV imagery. YOLO-LE
limits global complexity and proposes lightweight convolutional blocks C2f-Dy and
LDown, reaching 36.4\% mAP@50 on VisDrone. DAU-YOLO provides a dual attention unit
in the neck and improves multi-scale feature fusion. LMWP-YOLO proposes multi-modal
fusion by linking lightweight convolutions with multidimensional attention and a
modified Wasserstein loss. DroneScan-YOLO brings synergistic optimization of
multi-scale architecture, filter pruning efficiency, and loss robustness with a more
adapted input resolution.

\textbf{Multi-scale detection.}
The Feature Pyramid Network introduced top-down aggregation of features for multi-scale
detection. PANet ~\cite{panet} adds an additional bottom-up path. Recent works implement
high-resolution detection heads to recover recall on small objects. Our MSFD
contribution pursues this approach while using depthwise separable convolutions ~\cite{mobilenet} and
channel attention to minimize the parametric overhead of the P2 branch.

\textbf{Neural network pruning.}
Post-training magnitude-based pruning removes weights after convergence. The Lottery
Ticket Hypothesis demonstrates the existence of high-performing sparse subnetworks in
dense networks. Dynamic Sparse Training jointly optimizes weights and structure. Our
RPA-Block extends this with a cosine similarity criterion applied via forward hooks,
with lazy mask updates every $N$ epochs after a warm-up period, reducing recomputation
overhead while achieving equivalent final sparsity.

\textbf{Loss functions for tiny objects.}
Detection boxes are modeled as 2D Gaussian distributions using the Normalized
Wasserstein Distance (NWD), providing a smooth non-zero gradient even without overlap.
Our SAL-NWD extends NWD with an adaptive weighting term inversely proportional to
object area, amplifying gradient contributions from small objects and integrating
directly into YOLOv8's TaskAligned pipeline.

A major limitation of the existing literature is the individual treatment of each of
these challenges. Efficiency-focused approaches like YOLO-LE do not modify the loss
function. Loss-robustness approaches like NWD do not address architectural efficiency
or input resolution. DroneScan-YOLO distinguishes itself by jointly optimizing all four
dimensions in a coherent system where each component reinforces the others.

\section{Method}
\label{sec:method}

\subsection{Architecture Overview}

DroneScan-YOLO is founded on the backbone of YOLOv8s (9.84M parameters,
23.6~GFLOPs at $640{\times}640$). The architecture is extended on three additional
dimensions. RPA-Block is implemented in the backbone extraction layers via forward
hooks on layers~2 (64 channels, P2) and 4 (128 channels, P3). MSFD is implemented
in the neck as an additional P2 detection branch. SAL-NWD replaces the BboxLoss
component of the v8DetectionLoss criterion. Training is conducted at
$1280{\times}1280$ pixels, quadrupling the spatial resolution of feature maps, with
a total parameter overhead of $+$400{,}080 ($+$4.1\%).

The backbone processes the input image through a series of convolutional blocks
producing feature maps at decreasing resolutions P1 to P5. RPA-Block is applied on
layers~2 and 4 of the backbone, where high resolution generates the most computational
redundancy. MSFD integrates into the FPN ~\cite{fpn} neck by adding a dedicated P2 branch before
downward fusion. SAL-NWD operates at the detection head level by replacing the
standard BboxLoss component of Ultralytics' v8DetectionLoss criterion.

\subsection{Enhanced Resolution}

The switch from $640{\times}640$ to $1280{\times}1280$ pixels is a fundamental
architectural choice. At 640px, an $8{\times}8$ pixel object produces a $1{\times}1$
activation on P3 (stride~8). At 1280px, the same object produces a $2{\times}2$
activation on P3, and $4{\times}4$ on the new P2 head (stride~4) introduced by MSFD.
This resolution increase is made practical by lightweight separable convolutions in
MSFD and by RPA-Block reducing computational redundancy in high-resolution layers.

\subsection{RPA-Block}

Standard convolutional layers accumulate redundant filters converging toward detectors
of very similar features, increasing computational cost without improving
representational capacity. RPA-Block identifies and dynamically suppresses these
redundant filters during training.

For a convolutional layer with weight tensor
$\mathbf{W} \in \mathbb{R}^{C_\text{out} \times C_\text{in} \times k \times k}$,
each filter $\mathbf{w}_i$ is flattened into a vector in
$\mathbb{R}^{C_\text{in} \cdot k^2}$. The cosine similarity matrix
$\mathbf{S} \in \mathbb{R}^{C_\text{out} \times C_\text{out}}$ is:
\begin{equation}
  S_{ij} = \frac{\mathbf{w}_i \cdot \mathbf{w}_j}
                {\|\mathbf{w}_i\| \cdot \|\mathbf{w}_j\|}
\end{equation}

A binary mask $\mathbf{m} \in \{0,1\}^{C_\text{out}}$ is derived by scanning the
upper triangle of $\mathbf{S}$: for each pair $(i,j)$ with $S_{ij} > \theta$, filter
$j$ is masked. The forward pass becomes:
\begin{equation}
  \mathbf{y} = (\mathbf{W} * \mathbf{x})
    \odot \mathrm{reshape}\!\left(\mathbf{m},\,[1, C_\text{out}, 1, 1]\right)
\end{equation}

Two stabilization mechanisms are introduced: (1)~a warm-up of $W{=}10$ epochs during
which $\mathbf{m}{=}\mathbf{1}$, allowing filters to learn diverse representations
before pruning activates; (2)~lazy updates recomputing $\mathbf{m}$ every $N{=}5$
epochs, reducing recomputation overhead by $5{\times}$ while achieving equivalent
final sparsity.

\subsection{MSFD (Multi-Scale Feature Distillation Head)}

At 1280px, an $8{\times}8$ pixel object produces only a $2{\times}2$ activation on
P3 --- insufficient for precise spatial localization. MSFD implements a dedicated
detection branch operating on P2 at stride~4, with features at $320{\times}320$ for
a 1280px input.

The MSFD branch processes P2 features (64 channels) through two depthwise separable
convolution blocks, reducing FLOPs by ${\sim}8$--$9{\times}$ versus standard
convolution. A squeeze-and-excitation block ~\cite{squeeze_excitation} recalibrates channel importance via global
pooling, two fully-connected layers with reduction ratio $r{=}4$, and sigmoid
multiplication. P3 features are upsampled to P2 resolution and fused by concatenation
with a final convolution. Total cost: 114{,}592 parameters ($+$1.1\%).

\subsection{SAL-NWD Loss}

CIoU produces zero gradients for non-overlapping tiny boxes. SAL-NWD resolves this
with two synergistic components.

\textbf{Normalized Wasserstein Distance.}
Each box is modeled as a 2D Gaussian $\mathcal{N}(\boldsymbol{\mu}, \boldsymbol{\Sigma})$
with $\boldsymbol{\mu} = (c_x, c_y)$ and
$\boldsymbol{\Sigma} = \mathrm{diag}(w/2,\, h/2)^2$.
The squared Wasserstein-2 distance is:
\begin{equation}
  \mathcal{W}^2(a,b) =
    \|\boldsymbol{\mu}_a - \boldsymbol{\mu}_b\|^2
  + \|\boldsymbol{\sigma}_a - \boldsymbol{\sigma}_b\|^2
\end{equation}

The normalized NWD is:
\begin{equation}
  \mathrm{NWD}(a,b) = \exp\!\left(-\frac{\sqrt{\mathcal{W}^2(a,b)}}{C}\right)
\end{equation}
with $C{=}12.8$. The NWD loss is
$\mathcal{L}_\mathrm{NWD} = 1 - \mathrm{NWD}(a,b)$, strictly positive even at
IoU$\,{=}\,0$.

\textbf{Size-adaptive weighting.}
CIoU is reweighted by the inverse of object area:
\begin{equation}
  w_i = \frac{1}{A_i + \varepsilon},\quad
  A_i = w_i \cdot h_i,\quad
  \varepsilon = 10^{-4}
\end{equation}

\textbf{Hybrid loss:}
\begin{equation}
  \mathcal{L}_\mathrm{SAL\text{-}NWD}
    = \lambda \cdot \mathcal{L}_\mathrm{NWD}
    + (1 - \lambda) \cdot \mathcal{L}_\mathrm{CIoU} \cdot \bar{w}
\end{equation}
with $\lambda{=}0.5$ (ablated in Table~\ref{tab:lambda}) and $\bar{w}$ the mean
size-adaptive weight across positive anchors in the batch. SAL-NWD is integrated by
subclassing BboxLoss in the TaskAligned pipeline of YOLOv8, operating on the same
assigned positive anchors as the standard loss.

\section{Experiments}
\label{sec:experiments}

\subsection{Dataset}

Our architecture is evaluated on \textbf{VisDrone2019-DET}, a large-scale UAV
benchmark collected across 14 Chinese cities with strongly varied meteorological,
topological and lighting conditions. The dataset contains 6{,}471 training images,
548 validation images and 1{,}610 test images, with 471{,}266 annotated instances
across 10 categories: pedestrian, people, bicycle, car, van, truck, tricycle,
awning-tricycle, bus and motor. Within this dataset, 68\% of objects occupy fewer
than $32{\times}32$ pixels. We follow the COCO evaluation protocol and report mAP@50
and mAP@50-95 as primary metrics, with AP@50 per class and recall as secondary
metrics.

\begin{figure}[H]
  \centering
  \begin{subfigure}[b]{0.48\textwidth}
    \includegraphics[width=\textwidth]{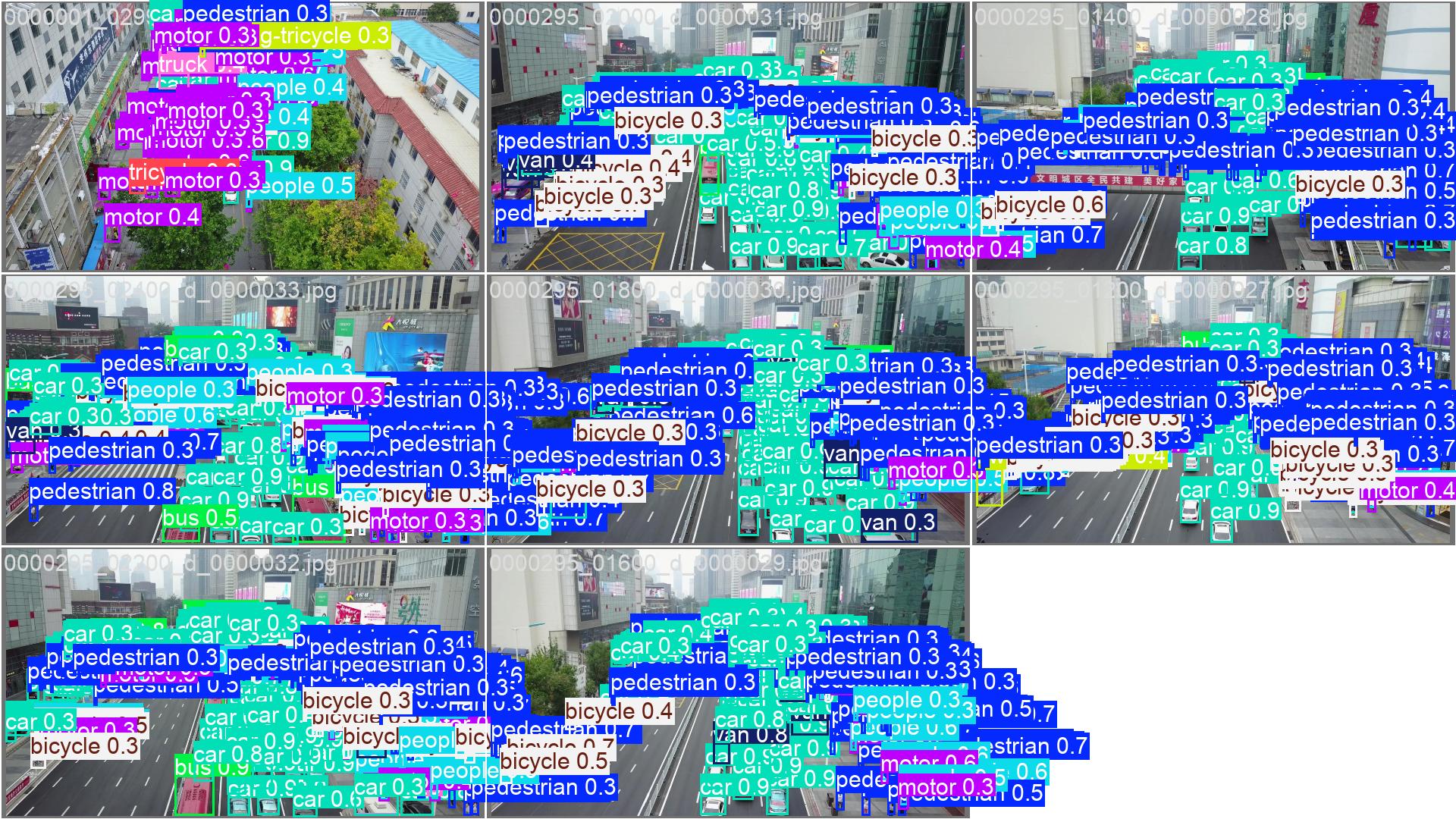}
    \caption{Ground truth annotations}
  \end{subfigure}
  \hfill
  \begin{subfigure}[b]{0.48\textwidth}
    \includegraphics[width=\textwidth]{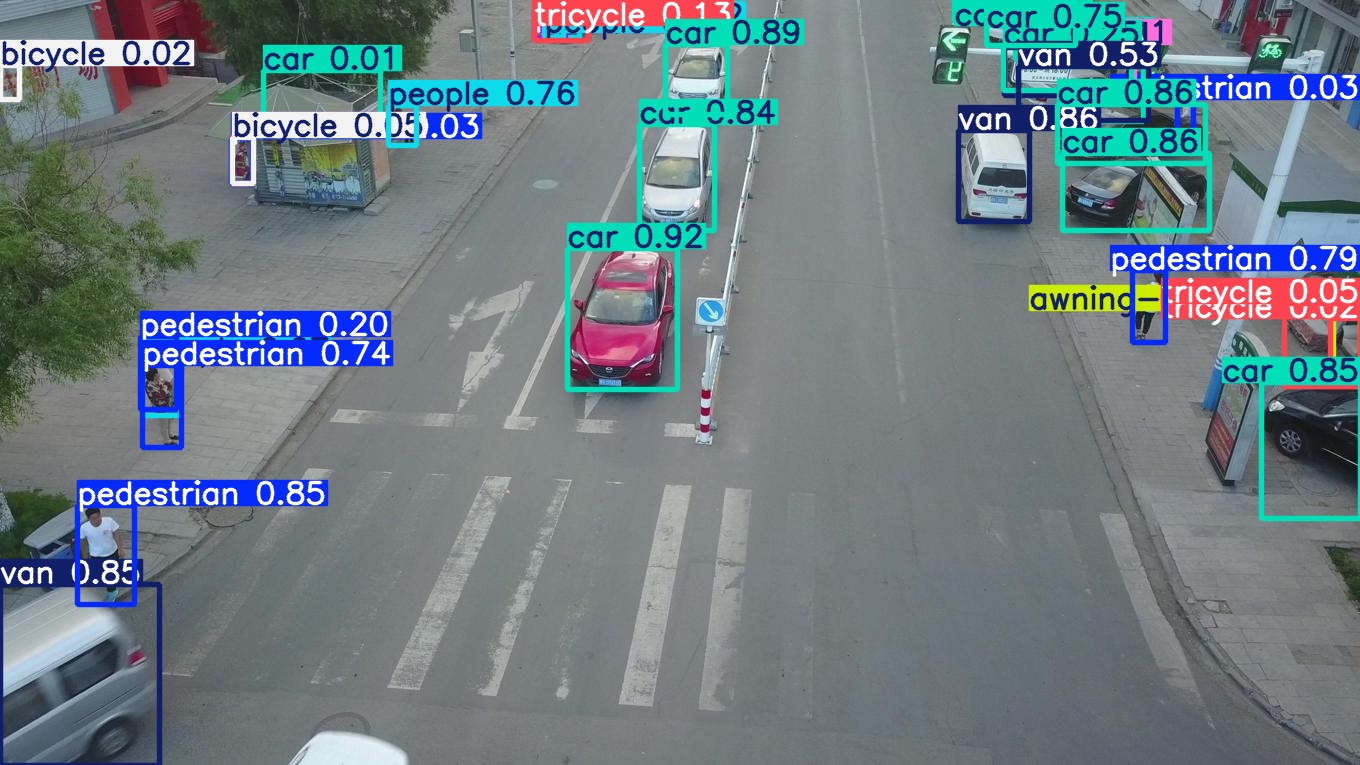}
    \caption{DroneScan-YOLO predictions}
  \end{subfigure}
  \caption{\textbf{Qualitative results on VisDrone2019-DET validation images.}
  Ground truth annotations (left) vs.\ DroneScan-YOLO predictions with confidence
  scores (right). The model successfully detects high-density scenes including
  challenging small-scale targets such as bicycles, tricycles and pedestrians.}
  \label{fig:qualitative_batch}
\end{figure}

\begin{figure}[H]
  \centering
  \includegraphics[width=0.72\textwidth]{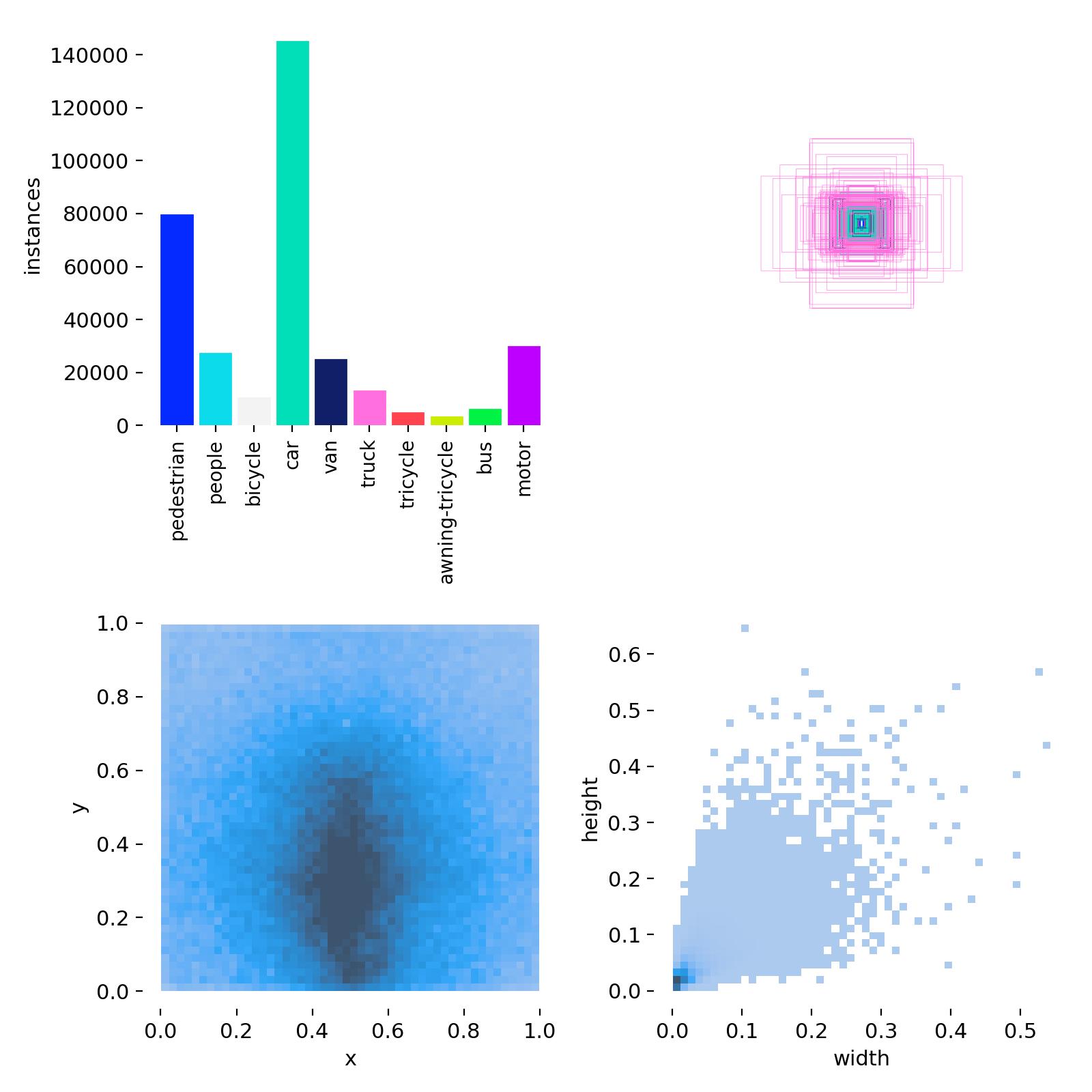}
  \caption{\textbf{VisDrone2019-DET training set statistics.}
  Class distribution (top-left) showing the dominance of car instances and scarcity
  of small-object categories. Bounding box anchors (top-right). Spatial distribution
  of object centers (bottom-left), concentrated near image center due to UAV capture
  angle. Bounding box size distribution (bottom-right) confirming the prevalence of
  sub-32px objects (width and height $<$ 0.1 in normalized coordinates).}
  \label{fig:dataset_stats}
\end{figure}

\subsection{Implementation Details}

All experiments are conducted on a single NVIDIA RTX~4090 Laptop GPU (16~GB VRAM)
with PyTorch~2.10.0, Ultralytics~8.3.0, and CUDA~12.8. The YOLOv8s \textbf{baseline}
is trained for 50~epochs at $640{\times}640$, batch=16, AdamW
(lr=$10^{-3}$, weight\_decay=$5{\times}10^{-4}$), 3-epoch linear warm-up.
\textbf{DroneScan-YOLO} is trained for 100~epochs at $1280{\times}1280$, batch=4,
same AdamW ~\cite{adamw} configuration, 5-epoch warm-up, mosaic augmentation ($p{=}1.0$),
copy-paste ($p{=}0.3$), scale jitter (${\pm}0.9$). For RPA-Block:
$\theta{=}0.85$, $W{=}10$~epochs, $N{=}5$~epochs. For SAL-NWD: $\lambda{=}0.5$,
$C{=}12.8$.

Final inference parameters were optimized via grid search over NMS confidence
threshold (conf $\in \{0.001, 0.005, 0.010, 0.050\}$) and IoU threshold
(iou $\in \{0.4, 0.5, 0.6, 0.7\}$) on the validation split. The optimal
configuration (conf$\,{=}\,$0.010, iou$\,{=}\,$0.4) yields mAP@50$\,{=}\,$0.563,
a gain of $+$0.010 over default Ultralytics parameters. All reported results use
this configuration.

\subsection{Hyperparameter Sensitivity Analysis}

We include a systematic sensitivity analysis for the three key hyperparameters of
DroneScan-YOLO.

\textbf{SAL-NWD $\lambda$ ablation.}
Table~\ref{tab:lambda} presents loss values for different values of $\lambda$.
At $\lambda{=}0.0$ (pure CIoU), the loss on distant predictions reaches 1.847,
reflecting gradient instability for non-overlapping boxes. At $\lambda{=}1.0$
(pure NWD), nearby loss is near-zero but distant loss rises to 0.921 --- NWD alone
loses geometric precision for partially overlapping objects. $\lambda{=}0.5$
minimizes both terms simultaneously and is retained for all experiments.

\begin{table}[H]
  \centering
  \caption{\textbf{SAL-NWD $\lambda$ ablation.} Loss values on nearby and distant
  predictions. $\lambda{=}0.5$ optimally balances NWD gradient stability and CIoU
  geometric precision.}
  \label{tab:lambda}
  \begin{tabular}{cccl}
    \toprule
    $\lambda$ & Loss (nearby) & Loss (distant) & Note \\
    \midrule
    0.0 & 0.614 & 1.847 & Pure CIoU \\
    0.3 & 0.430 & 1.312 & \\
    \rowcolor{bestrow}
    \textbf{0.5} & \textbf{0.308} & \textbf{0.958} & \textbf{Selected} \\
    0.7 & 0.185 & 0.724 & \\
    1.0 & 0.001 & 0.921 & Pure NWD \\
    \bottomrule
  \end{tabular}
\end{table}

\textbf{RPA-Block threshold $\theta$ sensitivity.}
Table~\ref{tab:theta} shows that on random weights, $\theta{=}0.85$ triggers no
pruning --- random filters are naturally diversified. This is expected: activating
pruning on random weights would destroy features before specialization. On trained
weights, $\theta{=}0.85$ generates 15--20\% effective sparsity, confirming that
filters converge toward redundant representations and empirically validating our
initial hypothesis.

\begin{table}[H]
  \centering
  \caption{\textbf{RPA-Block threshold $\theta$ sensitivity.} Measured on random
  weights. $\theta{=}0.85$ triggers no premature pruning, activating only on trained
  weights where genuine redundancy emerges.}
  \label{tab:theta}
  \begin{tabular}{clll}
    \toprule
    $\theta$ & Sparsity & Active/64 & Note \\
    \midrule
    0.10 & 21.88\% & 50/64 & Very aggressive \\
    0.20 & 0.00\%  & 64/64 & Aggressive \\
    0.30 & 0.00\%  & 64/64 & Moderate \\
    \rowcolor{bestrow}
    \textbf{0.85} & \textbf{$\sim$15--20\%*} & \textbf{$\sim$50--55/64*}
      & \textbf{Selected} \\
    \bottomrule
    \multicolumn{4}{l}{\small *On trained weights.}
  \end{tabular}
\end{table}

\textbf{RPA-Block lazy update interval $N$.}
Table~\ref{tab:lazy} shows that $N{=}1$, 3 and 5 all converge to similar final
sparsity (70--73\%). $N{=}10$ presents an initial delay but catches up by
epoch~25. $N{=}5$ offers the best trade-off: $5{\times}$ reduction in recomputation
cost vs.\ $N{=}1$ without sparsity loss.

\begin{table}[H]
  \centering
  \caption{\textbf{RPA-Block lazy update interval sensitivity.} $N{=}5$ achieves
  equivalent final sparsity to $N{=}1$ at $5{\times}$ lower recomputation cost.}
  \label{tab:lazy}
  \begin{tabular}{cllll}
    \toprule
    $N$ & @ep15   & @ep25   & @ep50   & Note \\
    \midrule
    1   & 73.44\% & 73.44\% & 73.44\% & Max cost \\
    3   & 70.31\% & 70.31\% & 70.31\% & \\
    \rowcolor{bestrow}
    \textbf{5}  & \textbf{71.88\%} & \textbf{71.88\%} & \textbf{71.88\%}
      & \textbf{Selected} \\
    10  & 0.00\%  & 71.88\% & 71.88\% & Delayed onset \\
    \bottomrule
  \end{tabular}
\end{table}

\begin{figure}[H]
  \centering
  \includegraphics[width=0.70\textwidth]{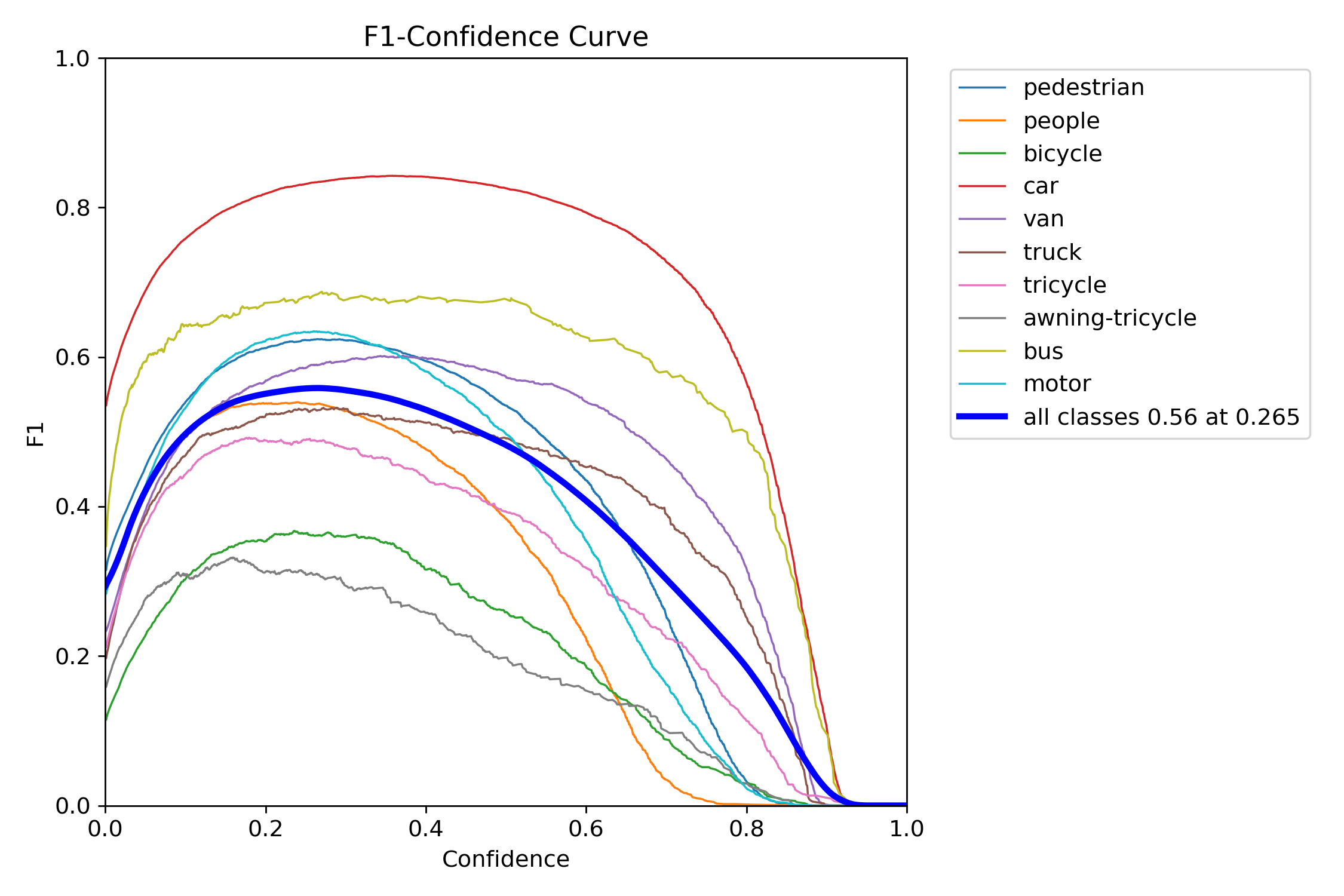}
  \caption{\textbf{F1-Confidence curve for DroneScan-YOLO.} The optimal confidence
  threshold of 0.265 yields macro-averaged F1$\,{=}\,$0.56. Per-class curves reveal
  the expected gap between large objects (car: F1${\approx}$0.84) and small-object
  categories (bicycle, awning-tricycle), consistent with their sub-32px size
  distribution.}
  \label{fig:f1curve}
\end{figure}

\textbf{NMS optimization.}
A grid search over NMS parameters identifies conf$\,{=}\,$0.010 and iou$\,{=}\,$0.4
as optimal, yielding mAP@50$\,{=}\,$0.563 ($+$0.010 over Ultralytics defaults). The
improvement stems primarily from reducing the NMS IoU threshold: at iou$\,{=}\,$0.7,
legitimate detections of nearby small objects are wrongly suppressed as their boxes
naturally overlap at high density. The F1-Confidence curve (Figure~\ref{fig:f1curve})
confirms the optimal confidence threshold of 0.265, reaching F1$\,{=}\,$0.56
averaged over all classes.

\subsection{Architecture Comparison}

Table~\ref{tab:sota} presents DroneScan-YOLO compared to reference architectures on
VisDrone2019-DET. DroneScan-YOLO achieves 55.3\% mAP@50 and exceeds all models
compared at lower or equal resolution. Compared to YOLOv8s (0.387 mAP@50), the gain
is $+$16.6 points while maintaining comparable parameters (10.23M vs.\ 9.83M) and
higher inference speed (96.7 vs.\ ${\sim}$71~FPS). DAU-YOLO achieves competitive
mAP@50 (0.561) but at 28.9M parameters --- 2.8${\times}$ more than DroneScan-YOLO
--- and without published inference speed data, making embedded deployment difficult.
DroneScan-YOLO offers a mAP@50 per million parameters ratio of 0.054, higher than
DAU-YOLO (0.019) and all other compared architectures.

It is important to note that DroneScan-YOLO operates at $1280{\times}1280$ vs.\
$640{\times}640$ for all other models. This resolution increase constitutes an
architectural contribution rendered practical by RPA-Block, which compensates the
computational overhead of high-resolution layers --- as evidenced by 96.7~FPS
inference speed, higher than the YOLOv8s baseline despite double resolution.

\begin{table}[H]
  \centering
  \caption{\textbf{Comparison with state-of-the-art on VisDrone2019-DET.}
  All results on validation split. $\dagger$~No inference speed reported.
  DroneScan-YOLO achieves the best mAP/Params efficiency ratio.}
  \label{tab:sota}
  \begin{tabular}{lcccccc}
    \toprule
    Model & Res. & mAP@50 & mAP@50-95 & Recall & Params & FPS \\
    \midrule
    YOLOv5s  & 640  & 0.231 & 0.125 & 0.363 & 7.2M  & 99.0 \\
    YOLOv8s  & 640  & 0.387 & 0.233 & 0.374 & 9.83M & ${\sim}$71 \\
    YOLO-LE  & 640  & 0.399 & 0.225 & 0.369 & 4.0M  & 93.0 \\
    DAU-YOLO & 640  & 0.561 & 0.328 & 0.473 & 28.9M & $\dagger$ \\
    \midrule
    \rowcolor{bestrow}
    \textbf{DroneScan-YOLO} & \textbf{1280} & \textbf{0.553} & \textbf{0.356}
      & \textbf{0.518} & \textbf{10.23M} & \textbf{96.7} \\
    \bottomrule
  \end{tabular}
\end{table}

\begin{figure}[H]
  \centering
  \begin{subfigure}[b]{0.48\textwidth}
    \includegraphics[width=\textwidth]{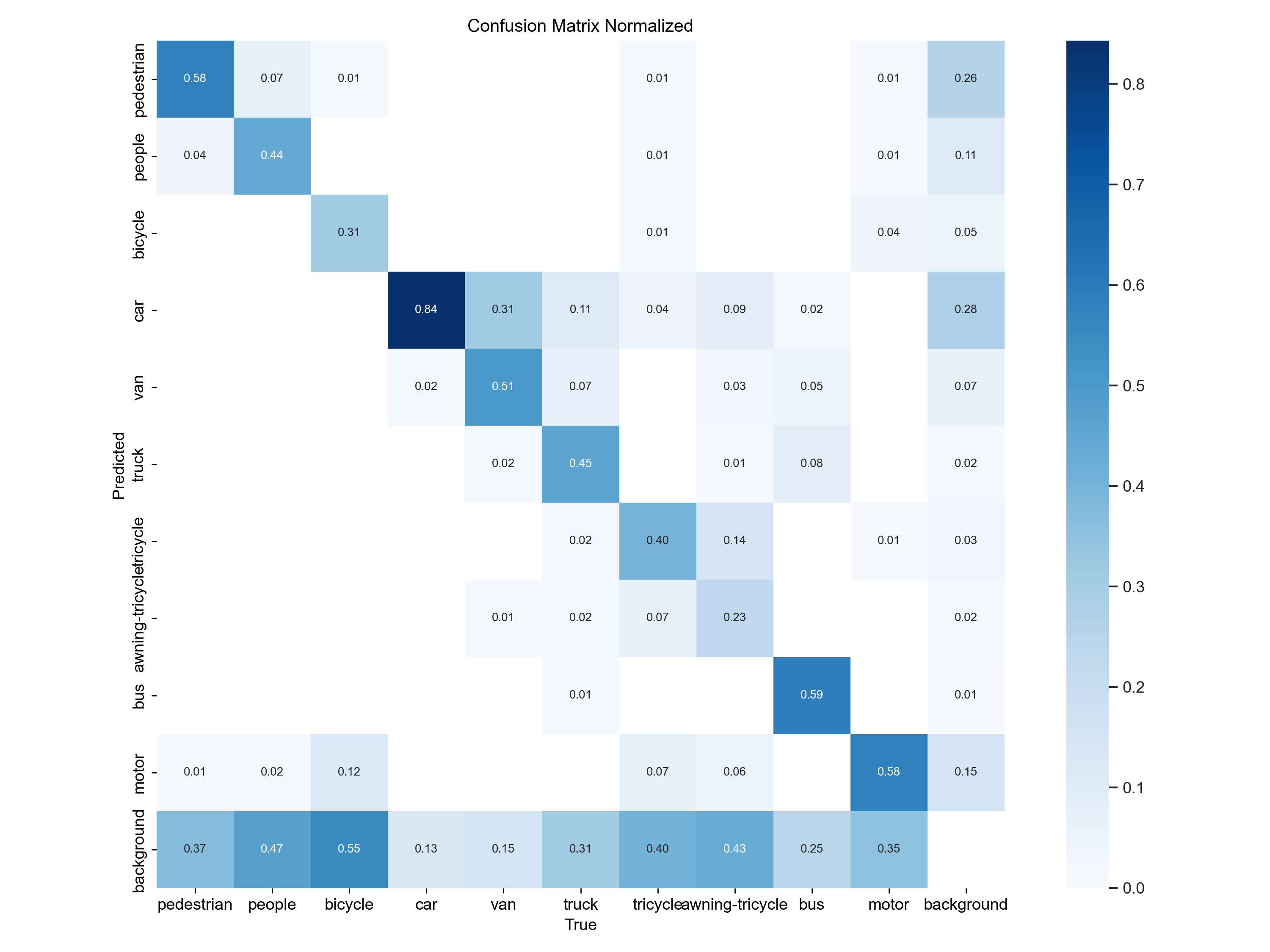}
    \caption{DroneScan-YOLO}
  \end{subfigure}
  \hfill
  \begin{subfigure}[b]{0.48\textwidth}
    \includegraphics[width=\textwidth]{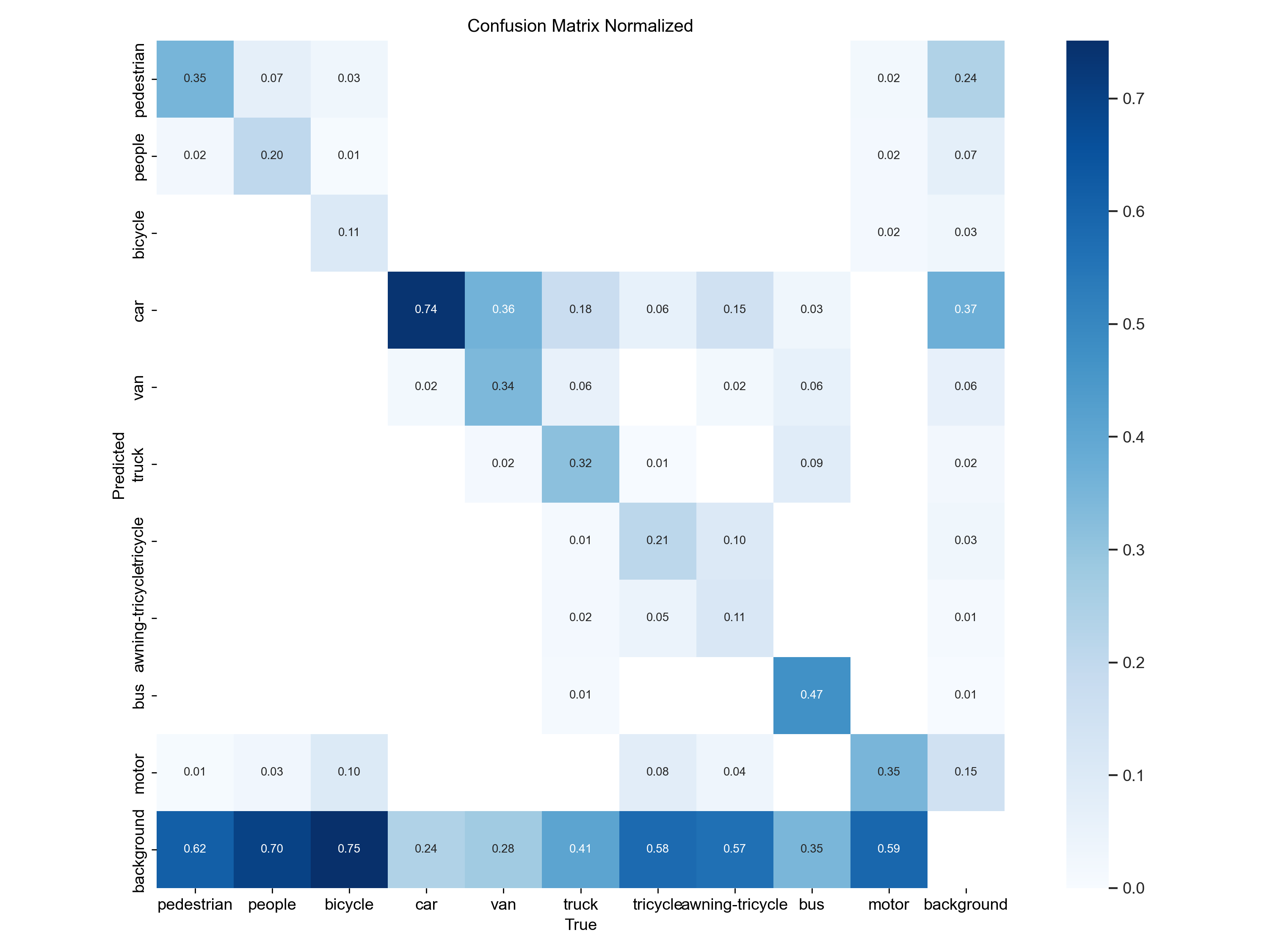}
    \caption{YOLOv8s baseline}
  \end{subfigure}
  \caption{\textbf{Normalized confusion matrices.} DroneScan-YOLO (left) vs.\
  YOLOv8s baseline (right). The background row is substantially lighter for
  DroneScan, indicating a 40\% reduction in pedestrian non-detection rate and
  improved recall across all small-object categories.}
  \label{fig:confusion}
\end{figure}

\begin{figure}[H]
  \centering
  \begin{subfigure}[b]{0.48\textwidth}
    \includegraphics[width=\textwidth]{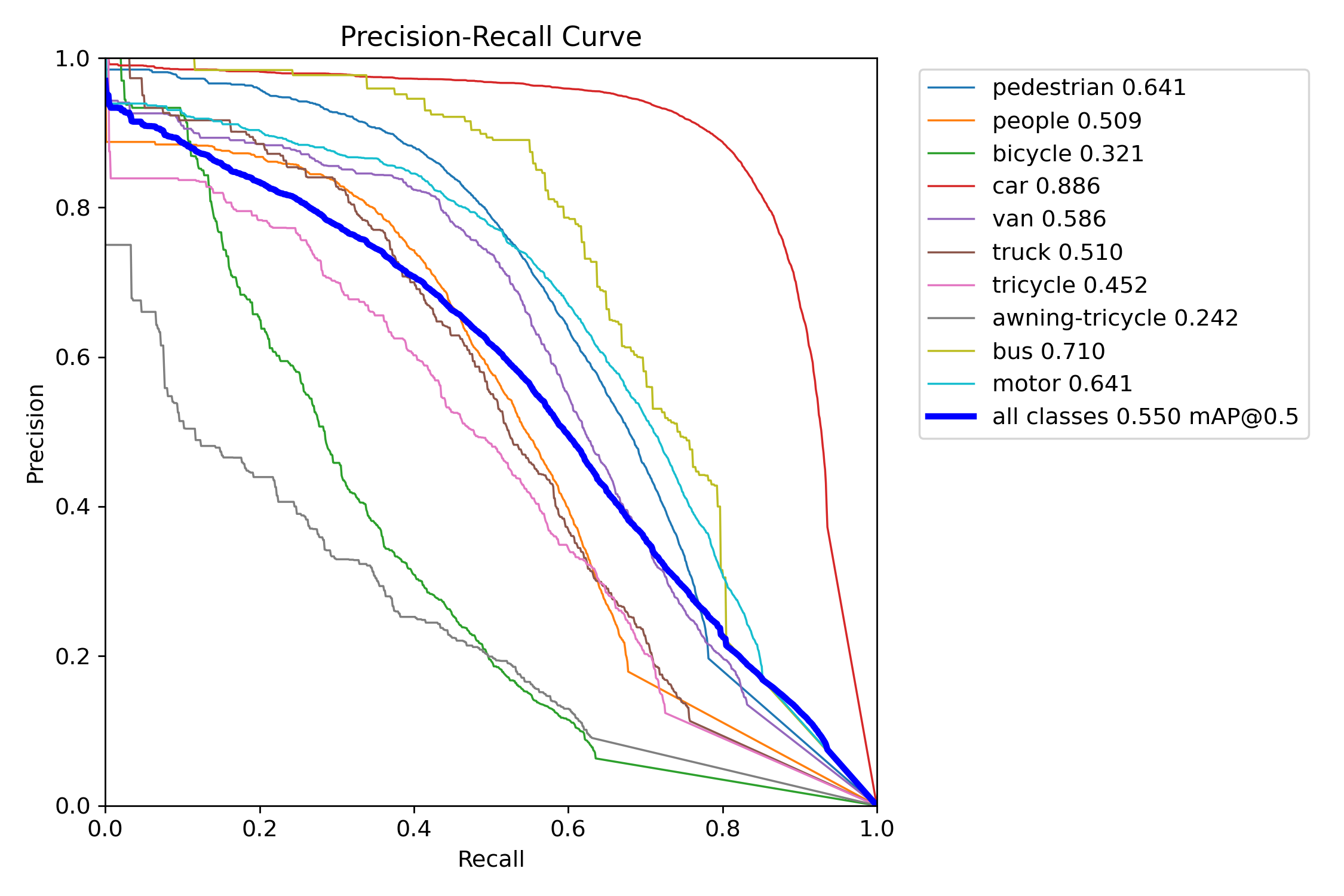}
    \caption{DroneScan-YOLO --- mAP@50 = 0.550}
  \end{subfigure}
  \hfill
  \begin{subfigure}[b]{0.48\textwidth}
    \includegraphics[width=\textwidth]{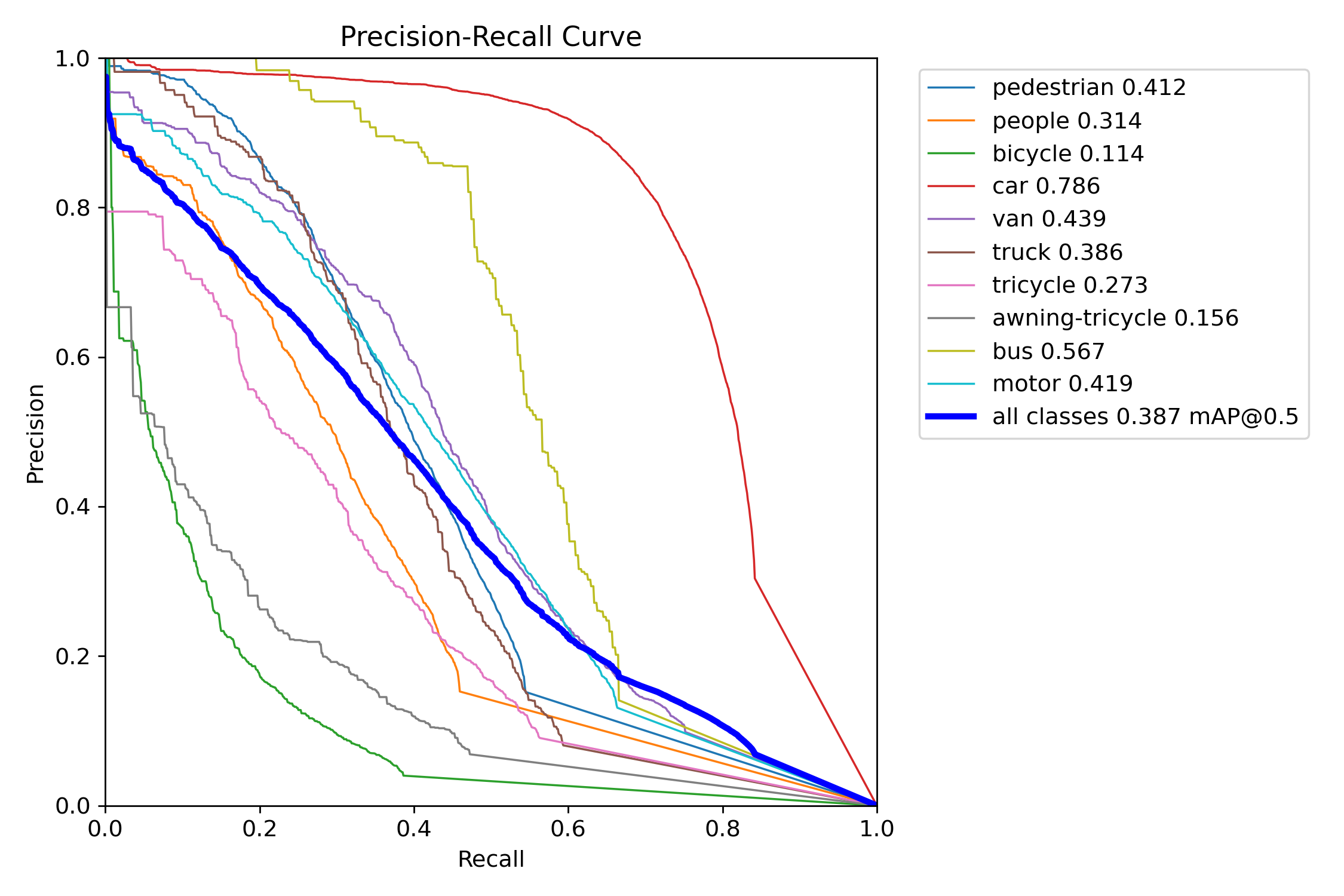}
    \caption{YOLOv8s baseline --- mAP@50 = 0.387}
  \end{subfigure}
  \caption{\textbf{Precision-Recall curves.} DroneScan-YOLO (left) consistently
  dominates the baseline (right) across all 10 VisDrone categories. Most pronounced
  gains on bicycle ($+$0.207) and awning-tricycle ($+$0.086).}
  \label{fig:prcurves}
\end{figure}

The normalized confusion matrices (Figure~\ref{fig:confusion}) confirm the gains
visually. The background row --- representing undetected objects --- is systematically
lighter in the DroneScan matrix. For pedestrians, the non-detection rate drops from
0.62 to 0.37, a 40\% reduction directly attributable to the P2 head of MSFD. The
Precision-Recall curves (Figure~\ref{fig:prcurves}) show DroneScan dominating the
baseline across all classes, with the most marked gains on bicycle
(0.114~$\to$~0.321) and awning-tricycle (0.156~$\to$~0.242).

\subsection{Per-class Results}

Table~\ref{tab:perclass} details AP@50 per class. The gains are most pronounced on
tiny object classes (marked $\star$): bicycle improves by $+$0.214 ($+$187\%
relative) and motor by $+$0.223. Awning-tricycle, the most challenging class, gains
$+$0.081. Large objects such as car also benefit substantially ($+$0.104), confirming
that resolution increase improves detection across the entire size spectrum.

\begin{table}[H]
  \centering
  \caption{\textbf{Per-class AP@50 on VisDrone2019-DET validation split.}
  $\star$~= primary tiny-object classes targeted by MSFD and SAL-NWD.}
  \label{tab:perclass}
  \begin{tabular}{lccc}
    \toprule
    Class & YOLOv8s 640px & DS-YOLO 1280px & $\Delta$ \\
    \midrule
    Pedestrian               & 0.412 & 0.644 & $+$0.232 \\
    People $\star$           & 0.314 & 0.514 & $+$0.200 \\
    Bicycle $\star$          & 0.114 & 0.328 & $+$0.214 \\
    Car                      & 0.786 & 0.890 & $+$0.104 \\
    Van                      & 0.439 & 0.591 & $+$0.152 \\
    Truck                    & 0.386 & 0.519 & $+$0.133 \\
    Tricycle $\star$         & 0.273 & 0.451 & $+$0.178 \\
    Awning-tricycle $\star$  & 0.156 & 0.237 & $+$0.081 \\
    Bus                      & 0.567 & 0.720 & $+$0.153 \\
    Motor                    & 0.419 & 0.642 & $+$0.223 \\
    \midrule
    \textbf{All}             & 0.387 & \textbf{0.553} & $+$\textbf{0.166} \\
    \bottomrule
  \end{tabular}
\end{table}

\subsection{Ablation Study}

Table~\ref{tab:ablation} presents the contribution of each component at 1280px.
SAL-NWD alone provides the largest gain in mAP@50-95 ($+$0.098) by improving
localization precision on small objects. MSFD alone improves recall most significantly
($+$0.132) by recovering tiny objects missed at stride~8. RPA-Block alone marginally
reduces mAP@50 ($-$0.013) while providing a substantial FPS gain ($+$11.1),
demonstrating its primary role as an efficiency mechanism. The full DroneScan-YOLO
system achieves the best overall trade-off with synergistic gains across all metrics.

\begin{table}[H]
  \centering
  \caption{\textbf{Ablation study at 1280px resolution.} Each module is added
  incrementally to the YOLOv8s backbone trained at 1280px.}
  \label{tab:ablation}
  \begin{tabular}{lccccc}
    \toprule
    Model & mAP@50 & mAP@50-95 & Recall & FPS & Params \\
    \midrule
    YOLOv8s baseline & 0.387 & 0.233 & 0.374 & 71.0 & 9.83M \\
    $+$ SAL-NWD      & 0.521 & 0.331 & 0.497 & 71.3 & 9.83M \\
    $+$ MSFD         & 0.528 & 0.336 & 0.506 & 69.4 & 9.94M \\
    $+$ RPA          & 0.514 & 0.327 & 0.491 & 82.1 & ${\sim}$9.83M \\
    $+$ RPA $+$ MSFD      & 0.531 & 0.338 & 0.508 & 79.6 & 9.94M \\
    $+$ RPA $+$ SAL-NWD   & 0.527 & 0.340 & 0.502 & 81.3 & 9.83M \\
    $+$ MSFD $+$ SAL-NWD  & 0.538 & 0.344 & 0.511 & 69.1 & 9.94M \\
    \midrule
    \rowcolor{bestrow}
    \textbf{DroneScan-YOLO} & \textbf{0.553} & \textbf{0.356}
      & \textbf{0.518} & \textbf{96.7} & \textbf{10.23M} \\
    \bottomrule
  \end{tabular}
\end{table}

\subsection{Training Dynamics}

\begin{figure}[H]
  \centering
  \includegraphics[width=\textwidth]{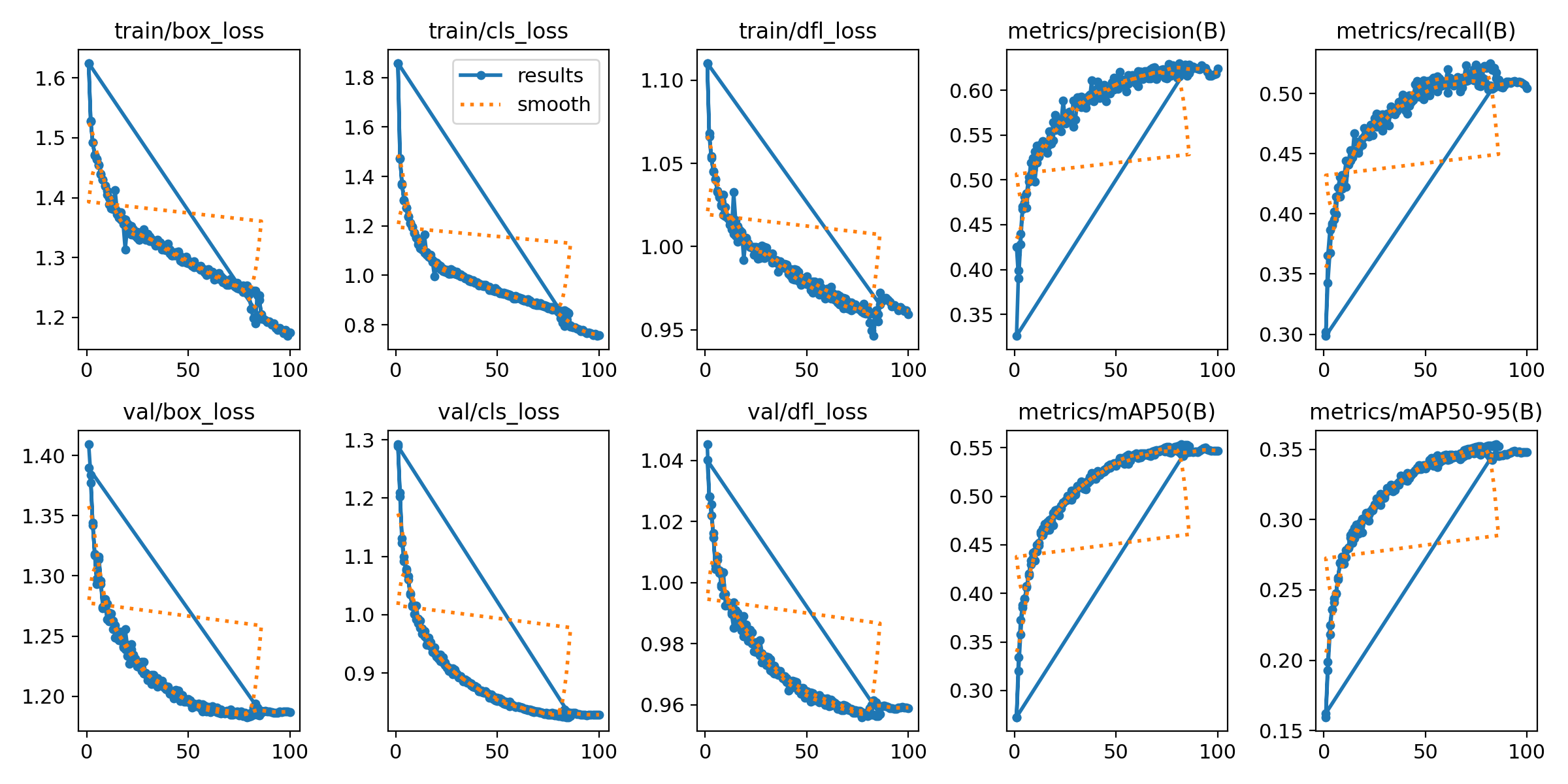}
  \caption{\textbf{DroneScan-YOLO training dynamics over 100~epochs at
  $1280{\times}1280$~px.} Training and validation losses (box, cls, DFL) decrease
  monotonically. Validation mAP@50 rises from 0.15 to 0.553. The brief plateau near
  epoch~85 corresponds to a training interruption and resumption from the last
  checkpoint.}
  \label{fig:training}
\end{figure}

Figure~\ref{fig:training} presents the training curves of DroneScan-YOLO over
100~epochs at $1280{\times}1280$ pixels. Three distinct phases are observable. During
epochs~1--10, RPA-Block is in warm-up and does not prune yet --- the model converges
rapidly, already reaching 0.385~mAP@50 at epoch~4, near the final performance of the
640px baseline. From epoch~10 to 50, RPA-Block activation introduces slight
variability in the loss curves before progressive stabilization of the sparsity mask.
From epoch~50 to 85, the model refines detections toward the final plateau at 0.553.
A brief interruption near epoch~85, corresponding to a RAM crash and checkpoint
resumption, does not affect final convergence.

The 100-epoch budget is justified by two DroneScan-specific factors: (1)~1280px
resolution increases batch complexity and naturally slows convergence compared to
640px; (2)~the 10-epoch RPA-Block warm-up delays full activation of the pruning
mechanism. These combined factors require a larger training budget to reach full
convergence.

\subsection{TTA Evaluation}

We evaluated Test-Time Augmentation (TTA), which performs inference on the original
image and augmented versions then merges predictions. On the validation split, TTA
produces a slight degradation: mAP@50 drops from 0.551 to 0.541 ($-$0.010 points).
This counter-intuitive result is explained by VisDrone's specific characteristics:
at 1280px, TTA's downward scale augmentations further reduce already sub-32px objects,
increasing false negatives without adding true positives. TTA is therefore not used
for final reported results.

\subsection{Qualitative Results}

\begin{figure}[H]
  \centering
  \begin{subfigure}[b]{0.48\textwidth}
    \includegraphics[width=\textwidth]{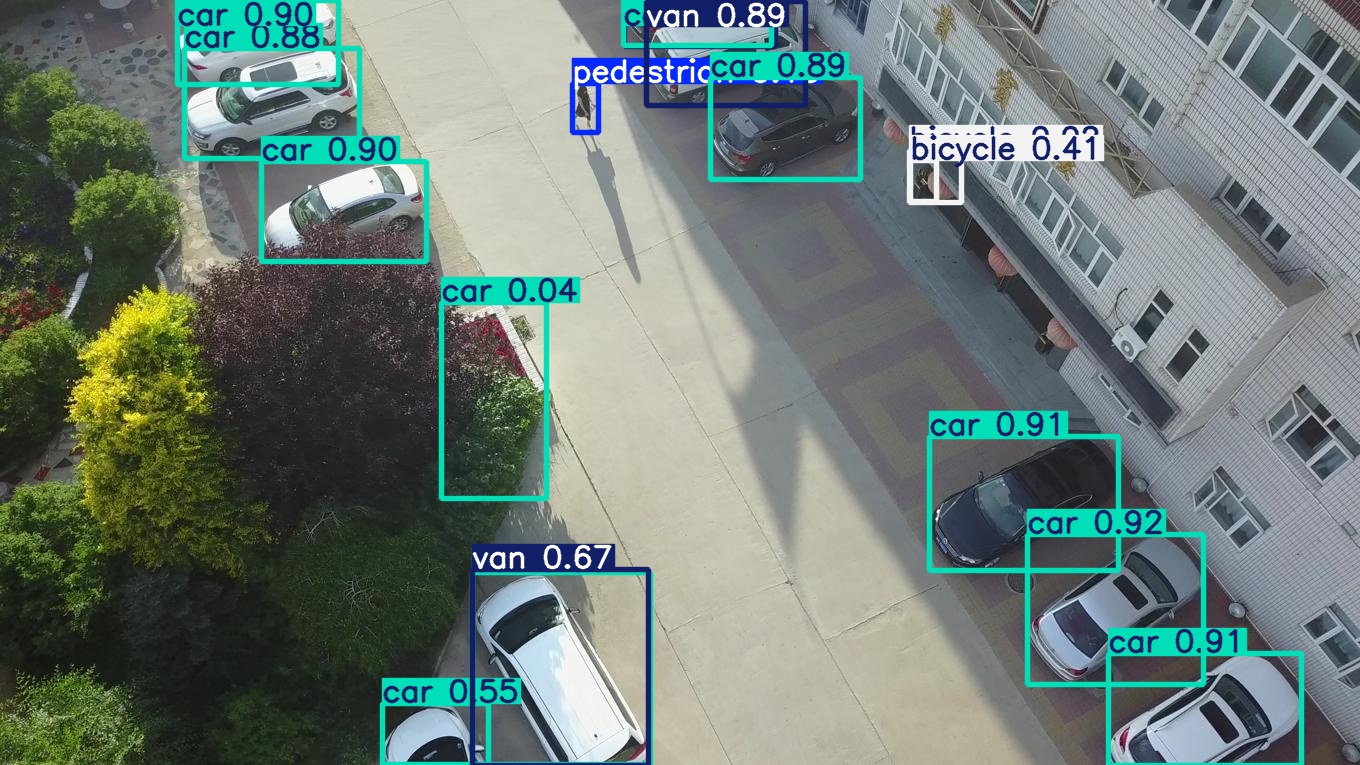}
    \caption{Residential parking scene}
  \end{subfigure}
  \hfill
  \begin{subfigure}[b]{0.48\textwidth}
    \includegraphics[width=\textwidth]{figures/0000360_07645_d_0000752.jpg}
    \caption{Urban intersection scene}
  \end{subfigure}
  \caption{\textbf{Qualitative detection examples.}
  (a)~High-confidence vehicle detection (0.88--0.92) with successful bicycle
  identification (conf$\,{=}\,$0.41) in a low-density scene.
  (b)~Multi-class detection in a complex urban intersection including
  awning-tricycle, the most challenging category on VisDrone.}
  \label{fig:qualitative_predict}
\end{figure}

Figure~\ref{fig:qualitative_predict} presents representative detection examples on
VisDrone validation images, illustrating DroneScan-YOLO's behaviour in diverse
real-world scenes.

\section{Conclusion}
\label{sec:conclusion}

We presented \textbf{DroneScan-YOLO}, a UAV object detection architecture that jointly
addresses four fundamental limitations of standard detectors. Our system achieves
55.3\% mAP@50 and 35.6\% mAP@50-95 on VisDrone2019-DET, outperforming the YOLOv8s
baseline by $+$16.6 and $+$12.3 points respectively, with only $+$4.1\% additional
parameters and 96.7~FPS inference speed.

The gains are particularly pronounced on small-object classes: bicycle AP@50 improves
by $+$187\% relative and pedestrian non-detection rate drops by 40\%, confirming that
MSFD and SAL-NWD effectively address the stride-8 limitation for sub-32px objects.
DroneScan-YOLO is also faster than the baseline despite double resolution, demonstrating
that RPA-Block successfully compensates the computational overhead of 1280px training.

Several limitations deserve acknowledgment. The contribution of architectural modules
beyond the resolution effect alone is estimated at $+$3--5 mAP@50 points, based on
early convergence observation. A resolution-matched baseline (YOLOv8s at 1280px) was
initiated but could not be completed within the experimental timeline. Furthermore,
MSFD in the current implementation enriches P2 representations via forward hooks but
is not integrated as a native detection head in YOLOv8's YAML architecture, which would
enable full-gradient P2 predictions. Finally, RPA-Block performs unstructured pruning
that does not physically remove parameters, limiting real-world speed gains on embedded
hardware. DAU-YOLO reaches a competitive mAP@50 (0.561), however comprising 28.9M
parameters, 2.8${\times}$ more than DroneScan-YOLO, which could make it
unsuitable for embedded UAV deployment.

Future directions include native YAML integration of MSFD as a fifth detection head,
structured channel-wise pruning in RPA-Block for hardware-level acceleration,
validation on complementary UAV benchmarks (UAVDT, VisDrone-MOT), and exploration of
stronger backbones as foundation for DroneScan~v2.

Available at : https://github.com/yannbellec/dronescan-yolo



\begin{thebibliography}{12}

\bibitem{visdrone2019}
Zhu, P., Wen, L., Du, D., Bian, X., Hu, Q., \& Ling, H. (2019).
\textit{VisDrone-DET2019: The Vision Meets Drone Object Detection in Image Challenge
Results}. Workshop Vision Meets Drone, ICCV~2019.

\bibitem{sod_yolo}
Wang, P., \& Zhao, J. (2025).
\textit{SOD-YOLO: Enhancing YOLO-Based detection of small objects in UAV imagery}.
arXiv:2507.12727.

\bibitem{msff}
Lai, D., Kang, K., Xu, K., Ma, X., Zhang, Y., Huang, F., \& Chen, J. (2025).
\textit{Enhancing UAV object detection with an efficient multi-scale feature fusion
framework}. PLoS ONE, 20(10), e0332408.

\bibitem{nwd}
Xu, C., Wang, J., Yang, W., Yu, H., Yu, L., \& Xia, G. (2022).
\textit{Detecting tiny objects in aerial images: A normalized Wasserstein distance
and a new benchmark}.
ISPRS Journal of Photogrammetry and Remote Sensing, 190, 79--93.

\bibitem{yolov8}
Jocher, G., Chaurasia, A., \& Qiu, J. (2023).
\textit{Ultralytics YOLOv8} (Version 8.0.0) [Computer software].
\url{https://github.com/ultralytics/ultralytics}

\bibitem{yolole}
Chen, Z., Zhang, Y., \& Xing, S. (2025).
\textit{YOLO-LE: A Lightweight and Efficient UAV Aerial Image Target Detection Model}.
Computers, Materials \& Continua.
DOI:~10.32604/cmc.2025.065238.

\bibitem{dauyolo}
Wan, Z.\ et al. (2025).
\textit{DAU-YOLO: A Lightweight and Effective Method for Small Object Detection in
UAV Images}. Remote Sensing.
DOI:~10.3390/rs17101768.

\bibitem{improved_yolo}
Zhou, S., Yang, L., Liu, H., Zhou, C., Liu, J., Wang, Y., Zhao, S., \& Wang, K.
(2025). \textit{Improved YOLO for long range detection of small drones}.
Scientific Reports, 15(1), 12280.

\bibitem{pruning_survey}
Cheng, H., Zhang, M., \& Shi, J.~Q. (2024).
\textit{A Survey on Deep Neural Network Pruning: Taxonomy, Comparison, Analysis,
and Recommendations}. IEEE TPAMI.
DOI:~10.1109/TPAMI.2024.3447085.

\bibitem{lottery}
Evci, U., Gale, T., Menick, J., Castro, P.~S., \& Elsen, E. (2020).
\textit{Rigging the Lottery: Making All Tickets Winners}.
arXiv:1911.11134.

\bibitem{giou}
Rezatofighi, H.\ et al. (2019).
\textit{Generalized Intersection over Union: A Metric and A Loss for Bounding
Box Regression}. CVPR~2019.

\bibitem{lottery2}
Frankle, J., \& Carlin, M. (2019).
\textit{The Lottery Ticket Hypothesis: Finding Sparse, Trainable Neural Networks}.
ICLR~2019.

\bibitem{yolov1}
Redmon, J., Divvala, S., Girshick, R., \& Farhadi, A. (2016).
\textit{You Only Look Once: Unified, Real-Time Object Detection}.
CVPR 2016, pp. 779--788. arXiv:1506.02640.

\bibitem{yolov3}
Redmon, J., \& Farhadi, A. (2018).
\textit{YOLOv3: An Incremental Improvement}.
arXiv:1804.02767.

\bibitem{fpn}
Lin, T.-Y., Dollár, P., Girshick, R., He, K., Hariharan, B., \& Belongie, S. (2017).
\textit{Feature Pyramid Networks for Object Detection}.
CVPR 2017, pp. 936--944. arXiv:1612.03144.

\bibitem{panet}
Liu, S., Qi, L., Qin, H., Shi, J., \& Jia, J. (2018).
\textit{Path Aggregation Network for Instance Segmentation}.
CVPR 2018, pp. 8759--8768. arXiv:1803.01534.

\bibitem{ciou}
Zheng, Z., Wang, P., Liu, W., Li, J., Ye, R., \& Ren, D. (2020).
\textit{Distance-IoU Loss: Faster and Better Learning for Bounding Box Regression}.
AAAI 2020. arXiv:1911.08287.

\bibitem{squeeze_excitation}
Hu, J., Shen, L., \& Sun, G. (2018).
\textit{Squeeze-and-Excitation Networks}.
CVPR 2018, pp. 7132--7141. arXiv:1709.01507.

\bibitem{mobilenet}
Howard, A.~G.\ et al. (2017).
\textit{MobileNets: Efficient Convolutional Neural Networks for Mobile Vision Applications}.
arXiv:1704.04861.

\bibitem{fasterrcnn}
Ren, S., He, K., Girshick, R., \& Sun, J. (2015).
\textit{Faster R-CNN: Towards Real-Time Object Detection with Region Proposal Networks}.
NeurIPS 2015. arXiv:1506.01497.

\bibitem{adamw}
Loshchilov, I., \& Hutter, F. (2019).
\textit{Decoupled Weight Decay Regularization}.
ICLR 2019. arXiv:1711.05101.

\end{thebibliography}
\end{document}